\documentclass{article}
\usepackage{ijcai22}

\usepackage{times}

\usepackage{soul}
\usepackage{url}
\usepackage[hidelinks]{hyperref}
\usepackage[utf8]{inputenc}
\usepackage[small]{caption}
\usepackage{graphicx}
\usepackage{amsmath}
\usepackage{booktabs}
\usepackage{bm,amssymb}
\usepackage{bbm}
\usepackage{multirow}
\usepackage{makecell}
\usepackage{fdsymbol}
\usepackage{verbatim}
\usepackage{hyperref}
\makeatletter
\def\UrlAlphabet{%
      \do\a\do\b\do\c\do\d\do\e\do\f\do\g\do\h\do\i\do\j%
      \do\k\do\l\do\m\do\n\do\o\do\p\do\q\do\r\do\s\do\t%
      \do\u\do\v\do\w\do\x\do\y\do\z\do\A\do\B\do\C\do\D%
      \do\E\do\F\do\G\do\H\do\I\do\J\do\K\do\L\do\M\do\N%
      \do\O\do\P\do\Q\do\R\do\S\do\T\do\U\do\V\do\W\do\X%
      \do\Y\do\Z}
\def\UrlDigits{\do\1\do\2\do\3\do\4\do\5\do\6\do\7\do\8\do\9\do\0}
\g@addto@macro{\UrlBreaks}{\UrlOrds}
\g@addto@macro{\UrlBreaks}{\UrlAlphabet}
\g@addto@macro{\UrlBreaks}{\UrlDigits}
\makeatother

\usepackage{paralist}

\usepackage{color} 
\newcommand{\wxt}[1]{{#1}}

\title{Language Models as Knowledge Embeddings}

\author{
Xintao Wang$^1$\and
Qianyu He$^1$\and
Jiaqing Liang$^1$\footnote{Corresponding Authors}\and
Yanghua Xiao$^{1,2*}$\\
\affiliations
$^1$Shanghai Key Laboratory of Data Science, School of Computer Science, Fudan University\\
$^2$Fudan-Aishu Cognitive Intelligence Joint Research Center\\
\emails
\{xtwang21, qyhe21\}@m.fudan.edu.cn,
l.j.q.light@gmail.com,
shawyh@fudan.edu.cn
}

\begin{document}

\maketitle

\begin{abstract}
Knowledge embeddings (KE) represent a knowledge graph (KG) by embedding entities and relations into continuous vector spaces.
Existing methods are mainly structure-based or description-based. 
Structure-based methods learn representations that preserve the inherent structure of KGs. 
They cannot well represent abundant long-tail entities in real-world KGs with limited structural information.
Description-based methods leverage textual information and language models. 
Prior approaches in this direction suffer from problems like expensive negative sampling and restrictive description demand.
In this paper, we propose LMKE, which adopts 
\textbf{L}anguage \textbf{M}odels to derive \textbf{K}nowledge \textbf{E}mbeddings, aiming at both enriching representations of long-tail entities and solving problems of prior description-based methods.
We formulate description-based KE learning with a contrastive learning framework to improve efficiency in  training and evaluation.
Experimental results show that LMKE achieves state-of-the-art performance on KE benchmarks of link prediction and triple classification, especially for long-tail entities.


\end{abstract}

\section{Introduction}

Knowledge graphs (KGs) are multi-relational graphs composed of entities as nodes and relations as edges, such as WordNet~\cite{miller1995wordnet}. 
They have been used to support a wide range of applications, including information retrieval, recommender systems and question answering. 
For better applications of symbolic knowledge in KGs in recent machine learning models, 
many efforts have been devoted to embedding KGs into low-dimension vector spaces, which are referred to as knowledge embeddings (KEs).
The principal applications of KEs are link prediction and triple classification in KGs, while there is also an increasing trend to use KEs in natural language processing (NLP) tasks like text generation.

\begin{figure}[t]
    \centering
        \includegraphics[width=1\linewidth]{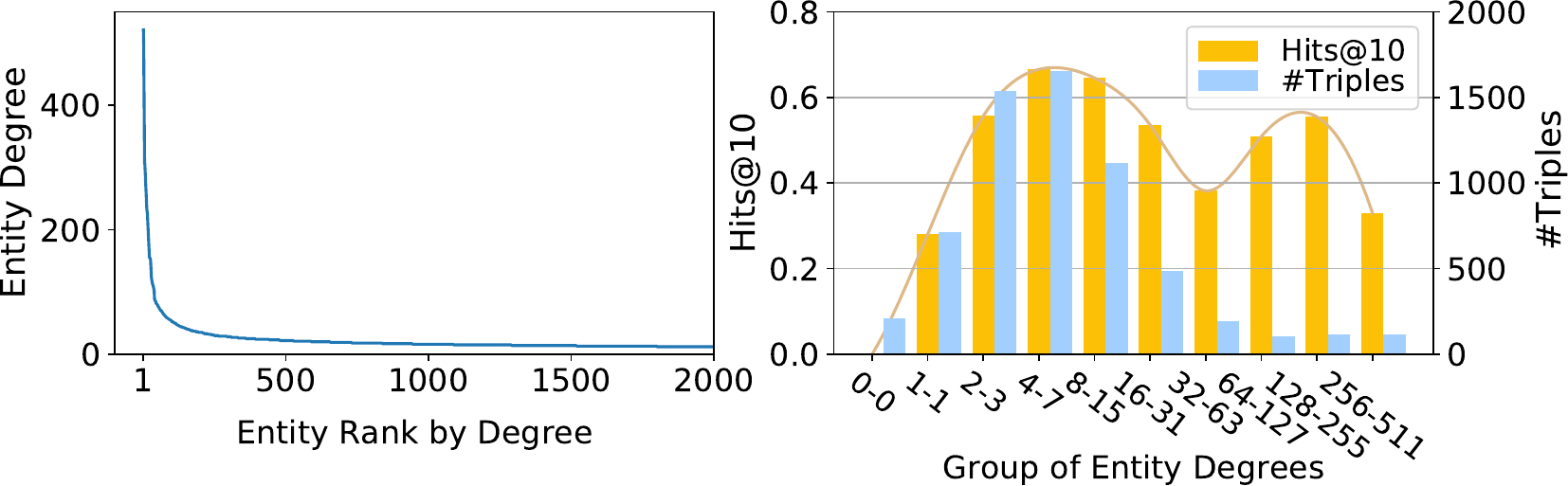}
    \captionsetup{font={small}} 
    \caption{
    Distribution of entity degrees and performance of RotatE (a typical structure-based method) on WN18RR. Entities are grouped by the logarithm of degrees (right). For each group, we report the number of relevant triples and their average performance on link prediction. Please refer to Sec 4.3 for more details.}
    \label{fig:long-tail}
\end{figure}

There are mainly two kinds of existing KE methods, namely 
traditional structure-based methods and emerging description-based methods.
Structure-based methods, such as TransE~\cite{bordes2013transe} and RotatE~\cite{sun2019rotate}, are trained to preserve the inherent structure of KGs. 
These methods cannot well represent long-tail entities, as they depend solely on the structure of KGs and thus favor entities rich in structure information (i.e., linked with plentiful entities).
However, real-world KGs are widely observed to have a right-skewed degree distribution, i.e., 
degrees of entities approximately follow the power-law distribution that forms a long tail, as is illustrated in Fig.\ref{fig:long-tail}. 
These KGs are populated with massive unpopular entities of low degrees.
For example, on WN18RR, 
14.1\% entities are with degree 1, 
and 60.7\% entities have no more than 3 neighbors.
Their embeddings consequently suffer from limited structural connectivity. 
This problem is justified by the declined performance of structure-based methods on long-tail entities shown in Fig.\ref{fig:long-tail}, which suggests that their embeddings are still unsatisfactory. 

Description-based KE methods represent entities in KGs by encoding their descriptions with language models, such as DKRL~\cite{xie2016dkrl} and KEPLER~\cite{wang2019kepler}. 
Textual descriptions provide rich information for numerous semantic-related tasks,   
which brings an opportunity to learn informative representations for long-tail entities. 
In an extreme case,  an emerging entity that is novel to an existing KG 
(in other words, having zero degrees in the KG) 
can still be well represented with its textual information.
This feature is referred to as a capacity in the ‘‘inductive (zero-shot) setting’’ by prior approaches.
Besides, lots of missing knowledge in KGs could be covered by rich textual information contained in entity descriptions or learned by pretrained language models (PLMs), given that current text corpora outstrip KGs in scale and contain much more information.  
However, existing description-based methods suffer from the following problems:

\begin{enumerate}
    \item \emph{Expensive negative sampling}. 
    While negative sampling is vital to KE learning, 
    existing description-based methods are only allowed to have limited negative samples
    because of the encoding expense of language models. 
    \item \emph{Restrictive description demand}. 
    Existing methods generally require descriptions of all entities in the KG, and discard entities with no or short descriptions. 
    Although delicate benchmarks can satisfy this demand, 
    real-world KGs can hardly contain descriptions for all entities.
    
\end{enumerate}



In this paper, we propose LMKE, which adopts \textbf{L}anguage \textbf{M}odels to derive  \textbf{K}nowledge \textbf{E}mbeddings,
aiming at both enhancing representations of long-tail entities and solving the above issues in prior description-based KEs.
LMKE leverages the inductive capacity of description-based methods to enrich the representations of long-tail entities.
In LMKE, entities and relations are regarded as special tokens  
whose output embeddings are contextualized in and learned from related entities, relations, and their textual information, 
so LMKE is also suitable for entities with no description.
A contrastive learning framework is further proposed 
where entity embeddings within a mini-batch serve as negative samples for each other 
to avoid the additional cost of encoding negative samples.   
In summary, our contributions are listed as follows:\footnote{Our codes are available at \url{https://github.com/Neph0s/LMKE}}
\begin{enumerate}
    \item We identify the problem of structure-based KEs in representing long-tail entities, and generalize the inductive capacity of description-based KEs to solve this problem. 
    To the best of our knowledge, we are the first to propose leveraging textual information and language models to enrich representations of long-tail entities.
    \item We propose a novel method LMKE that tackles two deficiencies of existing description-based methods, namely expensive negative sampling and restrictive description demand. 
    We are also the first to formulate description-based KE learning as a contrastive learning problem.
    \item We conduct extensive experiments on widely-used \quad KE benchmarks, 
    and show that LMKE achieves the state-of-the-art  performance on both link prediction and triple classification, surpassing existing structure-based and description-based methods, especially for long-tail entities. 
\end{enumerate}


\section{Related Work}
Knowledge embeddings represent entities and relations of KGs in low-dimension vector spaces.
KGs are composed of triples, where a triple $(h, r, t)$ means there is a relation $r\in\mathcal{R}$ between the head entity $h\in\mathcal{E}$ and the tail entity $t\in\mathcal{E}$. $\mathcal{E}$ and $\mathcal{R}$ denote the entity set and the relation set respectively. 

\paragraph{\textbf{Structure-Based Knowledge Embeddings.}}  
Existing KEs are mainly structure-based, deriving the embeddings by solely structure information of KGs.
These methods are further distinguished into translation-based models and semantic matching models in terms of their scoring functions~\cite{wang2017knowledge}. 
Translation-based models adopt distance-based scoring functions, which measure the plausibility of a triple $(h, r, t)$  by the distance between entity embeddings $\textbf{h}$ and $\textbf{t}$ after a translation specific to the relation.   
The most representative model is TransE. 
It embeds entities and relations as $\textbf{h}$, $\textbf{r}, \textbf{t}\in\mathbb{R}^{d}$ in a shared vector space with a dimension $d$.
Its loss function is defined as $\left\lVert \textbf{h}+\textbf{r}-\textbf{t} \right\rVert$ so as to make $\textbf{h}$ close to $\textbf{t}$ after a translation $\textbf{r}$. 
TransH~\cite{wang2014transh} proposes to project entity embeddings $\textbf{h}$ and $\textbf{t}$ to a relation-specific hyperplane, and TransR~\cite{lin2015transr} further proposes projections into relation-specific spaces.
RotatE defines a relation as a rotation from entity $h$ to entity $t$ in a complex vector space, so their embeddings $\textbf{h}$, $\textbf{r}$, $\textbf{t}\in\mathbb{C}^{d}$ are expected to satisfy $\textbf{h}\odot\textbf{r}\thickapprox\textbf{t}$, where $\odot$ stands for element-wise product.
Semantic matching models adopt similarity-based scoring functions, which measure the plausibility of a triple $(h, r, t)$ by matching latent semantics of $h, r, t$.
RESCAL~\cite{nickel2011rescal} represents the relation $r$ as a matrix $\textbf{M}_{r}$ and use a bilinear function $\textbf{h}^{\top}\textbf{M}_{r}\textbf{t}$ to score $(h, r, t)$. 
DistMult~\cite{yang2014distmult} makes $\textbf{M}_{r}$ diagonal for simplicity and efficiency. 
CoKE~\cite{wang2019coke} employs Transformers~\cite{vaswani2017attention} to derive contextualized embeddings, where triples or relation paths serve as the input token sequences. 

\begin{figure*}[htbp]
  \centering
  \includegraphics[width=0.9\linewidth]{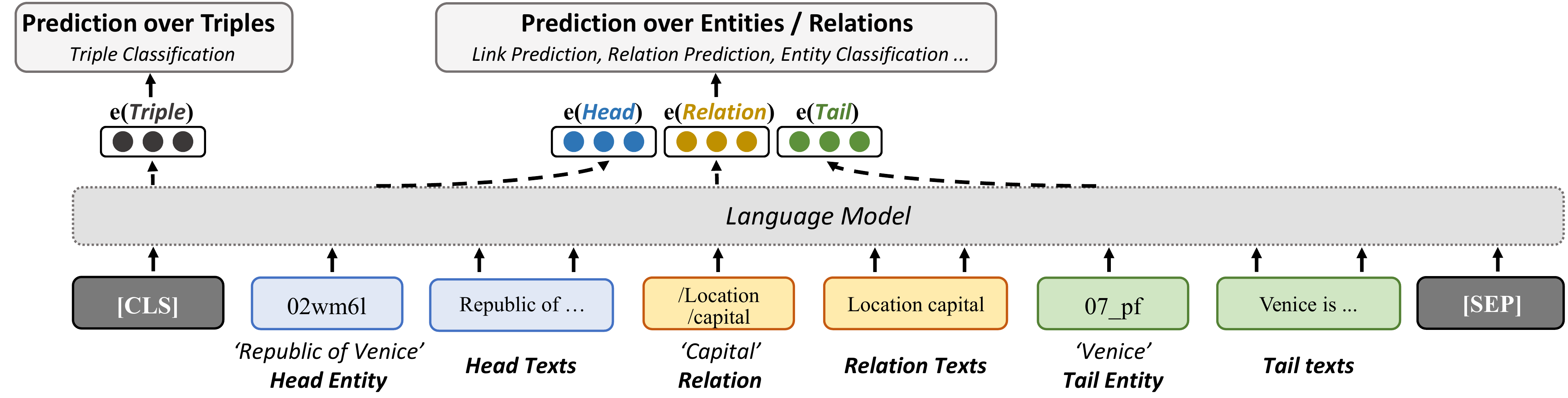}
  \caption{The architecture of LMKE.}
  \label{fig:triple_classification}
\end{figure*}

\paragraph{\textbf{Description-based Knowledge Embeddings.}}  
Recently, description-based KE methods are gaining increasing attention for the importance of textual information and development in NLP.
DKRL~\cite{xie2016dkrl} first introduces entities' descriptions and encodes them via a convolutional neural network for description-based embeddings. 
KEPLER~\cite{wang2019kepler} uses PLMs as the encoder to derive description-based embeddings and is trained with the objectives of both KE and PLM.
Pretrain-KGE~\cite{zhang2020pretrainkge} proposes a general description-based KE framework, which initializes another learnable KE with description-based embeddings and discards PLMs for efficiency after fine-tuning PLMs.
KG-BERT~\cite{yao2019kgbert} concatenates descriptions of $h$, $r$, $t$ as an input sequence to PLMs, and scores the triple by the sequence embedding.
StAR~\cite{wang2021star} thus partitions a triple into two asymmetric parts
between which it performs semantic matching. 


\section{Methods}
In this section, we introduce LMKE and its variants.
We first provide background on language models (Sec 3.1). 
Afterward, we elaborate on how LMKE adopts language models to derive knowledge embeddings (Sec 3.2), 
and its contrastive variants for both zero-cost negative sampling  in training and efficient link prediction in evaluation (Sec 3.3).

\subsection{Language Models}
Pretrained language models 
have been increasingly prevalent in NLP. 
They have been pretrained on large-scale corpora to store vast amounts of general knowledge. 
For example, BERT~\cite{devlin2018bert} is a Transformer encoder pretrained to predict randomly masked tokens.
Afterward, PLMs can be readily used to achieve excellent performance in various downstream tasks with fine-tuning. 

To better understand such excellence, knowledge probing is proposed~\cite{petroni2019lama}, which questions PLMs with masked cloze sentences. 
Researches in this direction have demonstrated that PLMs contain abundant factual knowledge and have the potential to be used as knowledge bases. 
Interestingly, PLMs are also shown to be capable of learning relations satisfying one-hop rules like equivalence, symmetry, inversion, and implication ~\cite{kassner-etal-2020-symbolic}, which is also the desiderata of knowledge embeddings.

\subsection{LMKE}

\begin{figure*}[htbp]
  \centering
   \includegraphics[width=0.9\linewidth]{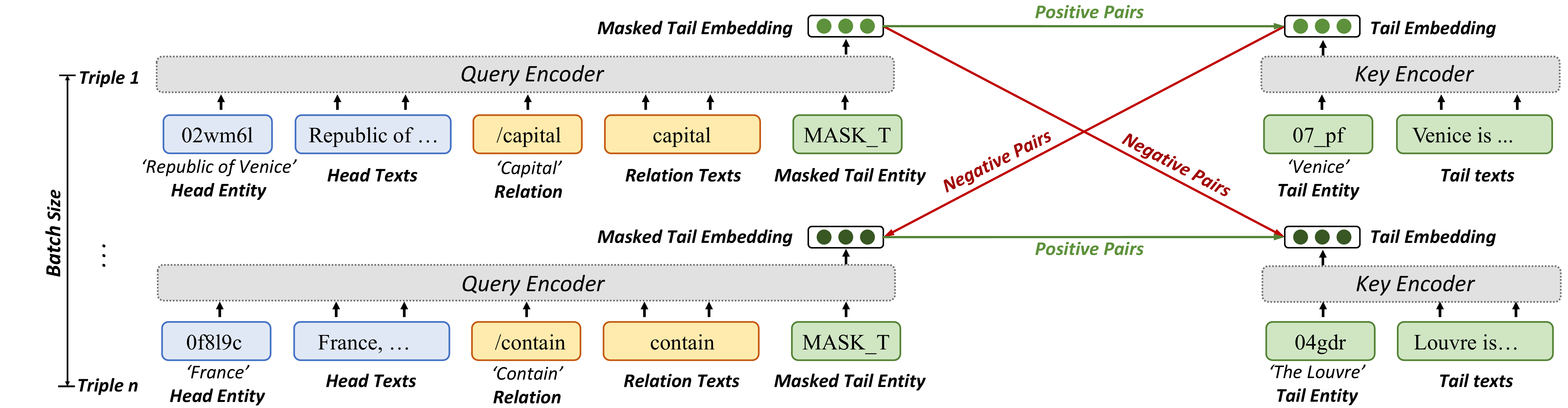}

  \caption{The architecture of contrastive LMKE. The queries and keys are encoded separately before contrastive matching within the batch. 
  }
  \label{fig:link_prediction_contrastive}

\end{figure*}



LMKE adopts language models as knowledge embeddings, i.e., deriving embeddings of entities and relations that can support prediction of the plausibility of given triples. 

LMKE learns embeddings of entities and relations in the same space as word tokens.
Given a specific triple $u=(h, r, t)$, LMKE leverages descriptions $s_{h}, s_{r}, s_{t}$ of $h, r, t$ as additional input, and outputs its probability $p(u)$.
LMKE makes every entity and relation a token and gets their embeddings via pretrained language models.
The triple is transformed into a sequence of tokens of $h, r, t$ and the words in their descriptions, which serves as the input to PLMs.
The description of an entity (or relation) $e$ is a sequence of tokens $s_{e}=(x_{1}, \ldots, x_{n_{e}})$ where $n_{e}$ denotes the length. 
The input sequence is thus 
$s_{u}=(h, s_{h}, r, s_{r}, t, s_{t})=(h, x_{1}^{h}, \ldots, x_{n_{h}}^{h}, r, x_{1}^{r}, \ldots, x_{n_{r}}^{r}, t, x_{1}^{t}, \ldots, x_{n_{t}}^{t})$.
Then, two special tokens [CLS] and [SEP] are inserted in the front and back of $s_{u}$.
Feeding forward $s_{u}$ into the PLM, 
LMKE generates the encoded embeddings $\textbf{h}, \textbf{r}, \textbf{t}\in\mathbb{R}^{d}$ of the three tokens $h, r, t$. 
In this way, we embed entities, relations, and words of their descriptions in a shared vector space. 
The embedding of one entity (or relation) is contextualized in and learned from not only its own textual information but also the other two components of the triple and their textual information.
Therefore, long-tail entities can be well represented with their descriptions, 
and entities without descriptions can also learn representations from related entities' descriptions. 
The output embedding of [CLS] aggregates information of the whole sequence, so we regard it as embedding $\textbf{u}$ for the entire triple. 
To score plausibility $p(u)$ of the triple, LMKE inputs $\textbf{u}$ into a linear layer like KG-BERT:
\begin{equation}
    p(u)= \sigma(\textbf{w}\textbf{u} + b)
\label{eq:cls}
\end{equation}
where $\textbf{w}\in\mathbb{R}^{d}$ and $b$ are learnable weight and bias. 
We adopt binary cross-entropy as the loss function for training. 
Besides,
LMKE can also be used for prediction over entities or relations with $\textbf{h}, \textbf{r}, \textbf{t}$. 
The architecture is shown in Fig. \ref{fig:triple_classification}. 


The principal applications of KEs are predicting missing links and classifying possible triples, which are formulated as two benchmark tasks for KE evaluation, namely \textbf{\textit{link prediction}}~\cite{bordes2013transe} and \textbf{\textit{triple classification}}~\cite{socher2013ntn}.
Triple classification is a binary classification task that judges whether a triple $u$ is true or not, which can be directly performed with  $p(u)$.
Link prediction is to predict the missing entity in a corrupted triple $(?, r, t)$ or $(h, r, ?)$ where $?$ denotes an entity removed for prediction. 
In this task, models need to corrupt a triple by replacing its head or tail entity with every entity in the KG, score the replaced triples, and rank the entities in descending order of the scores.
However, it is hardly affordable for LMKE to score every replaced triple as an integral with PLMs, concerning the time complexity shown in Table \ref{tab:complexity}. 
Even for a middle-sized dataset FB15k-237, there would be 254 million triples to be encoded. 


\begin{table}[htbp]
    \small
    \centering
    \scalebox{1}{
    \begin{tabular}{|c|c|c|}
    \Xhline{0.8pt}
        \textbf{Method} &
        \textbf{Training} &
        \textbf{Link Prediction}
        \\
    \Xhline{0.8pt}
        KG-BERT & $O(N_{\text{train}} N_{\text{neg}})$ & $O(|\mathcal{E}| N_{\text{eval}})$ \\ \hline
        StAR & $O(N_{\text{train}} N_{\text{neg}})$ & $O(|\mathcal{E}| |\mathcal{R}|)$ \\ \hline
        LMKE  & $O(N_{\text{train}} N_{\text{neg}})$ & $O(|\mathcal{E}| N_{\text{eval}})$ \\ \hline
        C-LMKE & $O(N_{\text{train}})$ & $O(N_{\text{eval}} + |\mathcal{E}|)$ \\
    \Xhline{0.8pt}
    \end{tabular}}
    \caption{The time complexity of training and link prediction evaluation. 
    $|\mathcal{E}|$ or $|\mathcal{R}|$ denotes the number of entities or relations in the KG. 
    $N_{\text{train}}$ or $N_{\text{eval}}$ is the number of triples in the training or evaluation split. $N_{\text{neg}}$ is the negative sampling size. C-LMKE denotes contrastive LMKE, whose complexity is lower than prior approaches.}
    \label{tab:complexity}
\end{table}


\subsection{Contrastive KE Learning}

Towards efficient link prediction with language models, 
one solution is to encode the triples partially.
Masked entity model (MEM-KGC)~\cite{choi2021memkgc} replaces the removed entity and its description with a mask $q$, and predicts the missing entity by feeding its output embedding $\textbf{q}$ into a linear layer.
It can be viewed as a masked variant of LMKE that trades off the exploitation of textual information for time complexity.
Since only one masked incomplete triple is encoded, the complexity is downgraded.
Nevertheless, it drops textual information of the target entities to be predicted, thus hurting the utility of textual information. 




We propose a contrastive learning framework to \wxt{better exploit} description-based KEs for link prediction,
where the \emph{given entity-relation pair} and the \emph{target entity} serve as the query $q$ and the key $k$ to be matched in contrastive learning. 


From this point of view, 
the output embedding $\textbf{q}$ of the masked entity in MEM-KGC is the encoded query, and the $i$-th row in the weights $\textbf{W}_{e}$ of the linear layer serves as the key that corresponds to the $i$-th entity for each $1\leq i\leq |\mathcal{E}|$.
Therefore, feeding $\textbf{q}$ into the linear layer can be viewed as matching the query $q$ with the keys. 
The difference is that the keys are directly learned representations instead of encodings of textual information like the queries. 

Contrastive LMKE (C-LMKE) is an LMKE variant under this framework that replaces the learned entity representations (rows of $\textbf{W}_{e}$) with encodings of target entities' descriptions, as is shown in Fig. \ref{fig:link_prediction_contrastive}. 
\wxt{It features \emph{the contrastive matching within the mini-batch}, which allows efficient link prediction without loss of information exploitation while avoiding the additional
cost of encoding negative samples.}
The candidate keys of a query are defined as keys of the queries in the batch.
C-LMKE's time complexity is analyzed in Table \ref{tab:complexity}.

This practice solves the problem of expensive negative sampling and allows description-based KEs to learn from much more negative samples.
While negative samples are vital to KE learning, most existing description-based methods are allowed to have only a few for each positive sample (usually ranging from 1 to 5) because of the cost of language models. 
C-LMKE currently binds the negative sampling size to the batch size,
and with our contrastive framework, existing approaches in contrastive learning like memory bank can be further introduced to achieve more improvement.

We match an encoded query $\textbf{q}$ with an encoded key $\textbf{k}$ by a two-layer MLP (multi-layer perceptron), instead of cosine similarity that is commonly adopted in contrastive learning, because there may exist multiple keys matching $q$.
If both $\textbf{k}_1$ and $\textbf{k}_2$ match $\textbf{q}$ and we maximize similarity between ($\textbf{q}$, $\textbf{k}_1$) and ($\textbf{q}$, $\textbf{k}_2$), ($\textbf{k}_1$, $\textbf{k}_2$) will also be enforced to be similar, which is not desirable. 
Thus, the probability that $q$ matches $k$ is:
\begin{equation}
    p(q, k)= \sigma(\text{MLP}[\textbf{q}; \textbf{k}; \textbf{q}-\textbf{k}; \textbf{q}\odot\textbf{k}; \textbf{d}])
\end{equation}
where  $\textbf{d}=[\log{(d_{q}+1)}; \log{(d_{k}+1)}]$ is the logarithm of entity degrees.
$d_{q}$ and $d_{k}$ are the degrees of the given entity in $q$ and the target entity $k$ counted on the training set. 
If the training set and the test set follow the same distribution, 
entities of higher degrees will also be more likely to appear in test triples,
so degrees are important structural information to be concerned.
While structure-based KEs learn degree information as aggregation into clusters~\cite{pei2019semi} and MEM-KGC learns it as the imbalance of entity labels, our matching function cannot capture this information, so we explicitly include it as additional features for prediction. 

Since there are usually multiple correct entities for a corrupted triple regarding relations that are not one-to-one, 
we adopt binary cross-entropy to judge whether each entity is a correct key (multi-label classification), instead of multi-class cross-entropy to find the most likely entity. 
Given that most of the keys are negative, we \wxt{average} losses over positive ones and negative ones respectively, and sum them up. 
\wxt{
The loss of matching a query $q$ with the keys $\mathcal{K}$ is thus formulated as 
$L(q, \mathcal{K}) = -\sum_{k^+\in\mathcal{K}^+} w_{q,k^+}\text{log}(p(q,k^+))-\sum_{k^-\in\mathcal{K}^-}w_{q,k^-}\text{log}(1-p(q,k^-))$, where $\mathcal{K}^+,\mathcal{K}^-$ denotes the positive and negative keys respectively and $\mathcal{K}=\mathcal{K}^+\cup\mathcal{K}^-$. 
We adopt self-adversarial negative sampling~\cite{sun2019rotate} for efficient KE learning, which calculates the weights as 
$w_{q,k^+}=\frac{1}{|\mathcal{K}^{+}|}$ and $w_{q,k^-}=\frac{\text{exp}(p(q,k^-))}{\sum_{k\in\mathcal{K}^-}\text{exp}(p(q,k))}$. In this case,  loss of false-positive samples is amplified and loss of true-negative samples is diminished.
}

\begin{table*}[htbp]
    \small
    \centering
    \resizebox{0.93\textwidth}{!}{
    \begin{tabular}{|c|c|c|c|c|c|c|c|c|c|c|}
    \Xhline{0.8pt}
        \textbf{Dataset} &
        \multicolumn{5}{c|}{\textbf{FB15k-237}} &
        \multicolumn{5}{c|}{\textbf{WN18RR}} 
        \\
    \hline
        \textbf{Metric} & 
        \textbf{MR} & \textbf{MRR} & \textbf{Hits@1} & \textbf{Hits@3} & \textbf{Hits@10} &
        \textbf{MR} & \textbf{MRR} & \textbf{Hits@1} & \textbf{Hits@3} & \textbf{Hits@10} \\
    \Xhline{0.8pt}
        \multicolumn{11}{|c|}{\textbf{Structure-based Knowledge Embeddings}} \\ \Xhline{0.8pt}
        $\text{TransE~\cite{bordes2013transe}}{\clubsuit}$ & 323 & 0.279 & 0.198 & 0.376 & 0.441 & 2300 & 0.243 & 0.043 & 0.441 & 0.532 \\ \hline
        $\text{DistMult~\cite{yang2014distmult}}{\spadesuit}$ & 254 & 0.241 & 0.155 & 0.263 & 0.419 & 5110 & 0.430 & 0.390 & 0.440 & 0.490 \\ \hline
        $\text{ComplEx~\cite{trouillon2016complex}}{\spadesuit}$ & 339 & 0.247 & 0.158 & 0.275 & 0.428 & 5261 & 0.440 & 0.410 & 0.460 & 0.510 \\ \hline
        $\text{RotatE~\cite{sun2019rotate}}{\spadesuit}$ & 177 & 0.338 & 0.241 & 0.375 & 0.533 & 3340 & 0.476 & 0.428 & 0.492 & 0.571 \\ \hline
        TuckER~\cite{balavzevic2019tucker} & - & 0.358 & 0.266 & 0.394 & 0.544 & - & 0.470 & 0.443 & 0.482 & 0.526 \\ \hline
        HAKE~\cite{zhang2020hake} & - & 0.346 & 0.250 & 0.381 & 0.542 & - & 0.497 & 0.452 & 0.516 & 0.582 \\ \hline
        CoKE~\cite{wang2019coke} & - & \textbf{0.364} & \textbf{0.272} & \textbf{0.400} & \textbf{0.549} & - & 0.484 & 0.450 & 0.496 & 0.553 \\ \Xhline{0.8pt}
        \multicolumn{11}{|c|}{\textbf{Description-based Knowledge Embeddings}} \\ \Xhline{0.8pt}
        $\text{Pretrain-KGE}_{\text{TransE}}$~\cite{zhang2020pretrainkge} & 162 & 0.332 & - & - & 0.529 & 1747 & 0.235 & - & - & 0.557 \\ \hline
        $\text{KG-BERT~\cite{yao2019kgbert}}{\clubsuit}$ & 153 & - & - & - & 0.420 & 97 & 0.216 & 0.041 & 0.302 & 0.524 \\ \hline
        $\text{StAR}_{\text{BERT-base}}\text{~\cite{wang2021star}}$ & \textbf{136} & 0.263 & 0.171 & 0.287 & 0.452 & 99 & 0.364 & 0.222 & 0.436 & 0.647 \\ \hline
        $\text{MEM-KGC}_{\text{BERT-base}}\text{(w/o EP)}$ & - & 0.339 & 0.249 & 0.372 & 0.522 & - & 0.533 & 0.473 & 0.570 & 0.636 \\ \hline
        $\text{MEM-KGC}_{\text{BERT-base}}\text{(w/ EP)}$ & - & 0.346 & 0.253 & 0.381 & 0.531 & - & 0.557 & 0.475 & 0.604 & 0.704 \\ \hline
        $\text{C-LMKE}_{\text{BERT-base}}$  & 141 & 0.306 & 0.218 & 0.331 & 0.484 & \textbf{79} & \textbf{0.619} & \textbf{0.523} & \textbf{0.671} & \textbf{0.789} \\ 
    \Xhline{0.8pt}
    \end{tabular}}
    \caption{Results of link prediction on FB15k-237 and WN18RR.
    ${\spadesuit}$ denotes results from ~\protect\cite{sun2019rotate}.
    ${\clubsuit}$ denotes results from ~\protect\cite{wang2021star}.
    We implement StAR on FB15k-237 with BERT-base as the base model.
    Other results are taken from their original papers.
    EP denotes the entity prediction task of MEM-KGC. 
    C-LMKE denotes contrastive LMKE.}
    \label{tab:main_exp}
\end{table*}


\section{Experiments and Analyses}

In this section, we evaluate the effectiveness of our methods. 

\subsection{Experiment Setup}


\paragraph{\textbf{Datasets.}}  
We experiment on four popular benchmark datasets: 
FB13~\cite{socher2013ntn}, 
FB15k-237~\cite{toutanova2015fb15k237}, UMLS~\cite{dettmers2018wn18rr} and WN18RR~\cite{dettmers2018wn18rr}, whose statistics are shown in Table \ref{tab:statistics}.
FB13 and FB15k-237 are derived from Freebase.
WN18RR is derived from WordNet. 
UMLS is a medical ontology describing relations between medical concepts. 
FB15k-237 and WN18RR are used for link prediction, where abundant inverse relations are removed in case they can serve as shortcuts. 
For triple classification, FB13 and UMLS are used.
Only the test split of FB13 contains negative samples.
In other situations, negative samples are created by randomly replacing head or tail entities, where the ground truth (training triples for the training split and all triples for the test split) would be avoided. 
Following KG-BERT, the entity descriptions we use are 
synsets definitions for WN18RR, and descriptions from Wikipedia for FB13, from~\cite{xie2016dkrl} for FB15k-237, and from~\cite{yao2019kgbert} for UMLS. 
The relation descriptions are their names for all datasets. 

\begin{table}[h!]
    \small
    \centering
    \setlength{\tabcolsep}{1mm}{
    \scalebox{0.9}{
    \begin{tabular}{*{7}{c}}
    \toprule
        \textbf{Dataset} & \textbf{\#Entity} & \textbf{\#Relation} & 
        \textbf{\#Train} & \textbf{\#Dev} & \textbf{\#Test} & \textbf{Avg DL}
        \\
        \midrule
    FB13 & 75,043 & 13 & 316,232 & 5,908 & 23,733 & 110.7 \\ 
    FB15k-237 & 14,541 & 237 & 272,115 & 17,535 & 20,466 & 141.7 \\ 
    UMLS & 135 & 46 & 5,216 & 652 & 661 & 161.7 \\ 
    WN18RR & 40,943 & 11 & 86,835 & 3,034 & 3,134 & 14.4 \\ 
    \bottomrule
    \end{tabular}}}
    \caption{Statistics of the datasets. \wxt{Avg DL means the average length (number of words) of descriptions.}}
    \label{tab:statistics}
\end{table}

\paragraph{\textbf{Baselines.}}  
We compare our methods with structure-based and description-based methods.
The structure-based methods include TransE, 
TransH, 
TransR, 
DistMult, 
ComplEx, 
RotatE. 
The description-based methods include 
Pretrain-KGE, 
KG-BERT, 
exBERT, 
and StAR. 
For a fair comparison, we reimplement StAR on FB15k-237 with BERT-base.

\paragraph{\textbf{Metrics.}}  
For triple classification, we report accuracy.
For link prediction, we report Mean Rank (MR), Mean Reciprocal Rank (MRR), and Hits@\{1, 3, 10\} in the ‘‘filtered’’ setting. 
Metrics for link prediction are based on the rank of the correct entity in a list of all entities ranked by their plausibility. 
The ‘‘filtered’’ setting is a common practice that removes other correct entities (which also constitute triples existing in the KG) from the list.
Hits@K measures the proportion of correct entities ranked in the top K.
The results are averaged over test triples and over predicting missing head and tail entities. 
Generally, a good model is expected to achieve higher MRR, Hits@N, and lower MR. 

\paragraph{\textbf{Settings.}}
We evaluate LMKE on triple classification and C-LMKE on link prediction
with BERT-base~\cite{devlin2018bert} 
as the language model. 
Larger models are not considered, in which case we have to decrease the batch size.
With BERT-base, we set the learning rate of PLMs and  other components as 10$^{-5}$  and 5$\times$10$^{-4}$ respectively. 
For triple classification, we set the batch size as 16 and sample 1 negative triple for each positive triple. 
For link prediction, we set the batch size as 256, and maximum number of tokens for each entity or relation as 50. We encode given entity-relation pair queries and target entity keys with two language models. They are equally initialized and share the same embeddings of words, entities, and relations.
Our models are fine-tuned with Adam as the optimizer. 
We select the best epoch based on accuracy or MRR on the dev set.  

\subsection{General Performance}

We compare our methods with prior approaches on both link prediction and triple classification.
Experimental results in Table \ref{tab:main_exp} and Table \ref{tab:main_exp_cls}  show that our methods achieve the state-of-the-art performance on both tasks.

Results of link prediction on WN18RR are demonstrated in Table \ref{tab:main_exp}, which show that 
our method significantly outperforms existing approaches.
C-LMKE with BERT-base achieves surpassing performance on MR, MRR and Hits@\{1, 3, 10\}.
The improvement is significant on WN18RR, where Hits@10 of our method reaches to 0.789 while the best result of previous methods is 0.704 from MEM-KGC. 
However, its entity prediction task is compatible with our work, without which its Hits@10 declines to 0.636. 
Nevertheless, our method still underperforms state-of-the-art structure-based methods on FB15k-237.

The results suggest that description-based methods largely outperform structure-based ones on WN18RR, but fail to surpass them on FB15k-237.  
We analyze the difference between these datasets to explain this phenomenon.
FB15k-237 differs from WN18RR mainly in two aspects: sparsity and description. 
According to statistics shown in Table \ref{tab:statistics}, 
the average degree on FB15k-237 and WN18RR is 37.4 and 4.2 respectively. 
The former is about 8.9 times the latter, which indicates that entities in FB15k-237 generally have access to rich structural information while entities in WN18RR are more likely to be long-tail. 
Also, textual information better covers structural information on WN18RR compared with FB15k-237. 
Entities on WN18RR are words, and descriptions are exactly their definitions from which structural relations are derived, so descriptions can ideally support the inference of structural relations.
However, entities on FB15k-237 are real-world entities whose collected descriptions only partially support the inference. 
For example, the fact (\textit{Albert Einstein, isA, peace activist}) is not covered by collected descriptions of these entities. 
As a result, the overreliance on textual information may hurt performance, and description-based methods did not achieve surpassing performance. 

\begin{table}[htbp]
    \small
    \centering
    \scalebox{0.9}{
    \begin{tabular}{|c|c|c|}
    \Xhline{0.8pt}
        \textbf{Dataset} &
        \textbf{FB13} &
        \textbf{UMLS}
        \\
    \Xhline{0.8pt}
        $\text{TransE~\cite{bordes2013transe}}$ & 81.5 & 78.1 \\ \hline
        $\text{TransH~\cite{wang2014transh}}$ & 83.3 & 79.2 \\ \hline
        $\text{TransR~\cite{lin2015transr}}$ & 82.5 & 81.9 \\ \hline
        $\text{DistMult~\cite{yang2014distmult}}$ & 86.2 & 86.8\\  \Xhline{0.8pt}
        $\text{KG-BERT~\cite{yao2019kgbert}}$ & 90.4 & 89.7 \\ \hline
        $\text{exBERT~\cite{yaser2021triple}}$ & - & 90.3 \\ \hline
        $\text{LMKE}_{\text{BERT-base}}$  & \textbf{91.7} & \textbf{92.4} \\ 
    \Xhline{0.8pt}
    \end{tabular}}
    \caption{Accuracy of triple classification on FB13 and UMLS.
    Results of existing baselines on FB13 and UMLS are taken from ~\protect\cite{yao2019kgbert} and ~\protect\cite{yaser2021triple} respectively. }
    \label{tab:main_exp_cls}
\end{table}

Results of triple classification on FB13 and UMLS in Table \ref{tab:main_exp_cls} show that LMKE also outperforms existing methods on this task. Comparison between results of LMKE and KG-BERT demonstrates the effectiveness of learning embeddings of entity and relation tokens in the same space as word tokens.

\subsection{Performance Grouped by Entity Degrees}
\begin{figure}[b]
    \centering
        \includegraphics[width=1\linewidth]{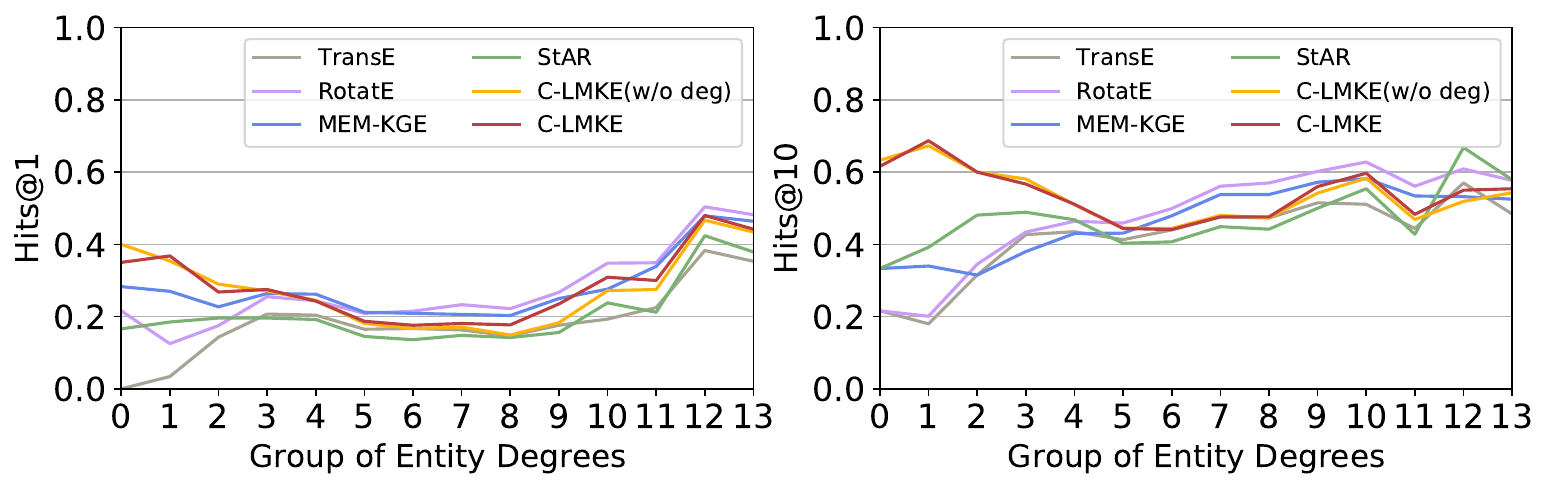}
    \captionsetup{font={small}} 
    \caption{
    Average Hits@1 and Hits@10 performance grouped by the logarithm of entity degrees on FB15k-237. }
    \label{fig:groupby}
\end{figure}
To show the effectiveness of our methods on long-tail entities, 
we group entities by the logarithm of degrees,
collect relevant triples for each group,
and study the average link prediction performance of different methods on different groups. 
A triple ($h$, $r$, $t$) is relevant to group $i$ if $h$ or $t$ is in group $i$. 
The grouping rule is the same as in Fig.\ref{fig:long-tail}.
The results on FB15k-237 in Fig.\ref{fig:groupby} show that description-based methods significantly outperform structure-based ones on long-tail entities in group 0, 1, 2, 3 and 4 (whose degrees are lower than 16), and our C-LMKE significantly surpasses other description-
based ones.  
Comparison between results of C-LMKE with or without degree information indicates that introducing degree information generally improves its performance on entities that are not long-tail.  
On popular entities, however, structure-based methods generally perform better. 
Although StAR is also description-based, it achieves the best Hits@10 on group 12 and 13, 
because it is trained with an additional objective that follows structure-based methods.

\subsection{Importance of Negative Sampling Size}

\begin{figure}[htbp]
    \centering
        \includegraphics[width=0.6\linewidth]{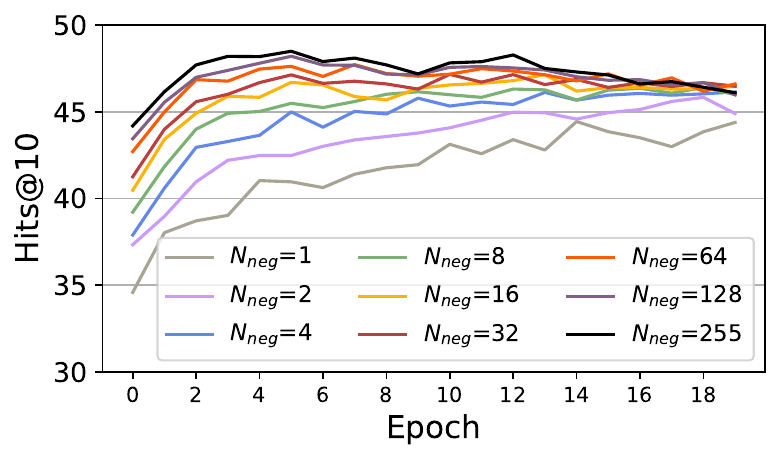}
    \captionsetup{font={small}} 
    \caption{Hits@10 of C-LMKE 
    on FB15k-237 with different negative sampling size. }\label{fig:nss}
\end{figure}

We study the performance of C-LMKE with different negative sampling sizes $N_{\text{neg}}$ to justify its importance. 
We set the batch size as 256 and limit $N_{\text{neg}}$ for each triple by only using several encoded target entities of other triples in the batch as negative keys. 
We report Hits@10 of C-LMKE 
on FB15k-237 for 20 epochs. 
The results shown in Fig. \ref{fig:nss} indicate that larger $N_{\text{neg}}$ continuously brings better performance and greatly accelerates training, especially when $N_{\text{neg}}$ is under 64. 
However, prior description-based methods generally set $N_{\text{neg}}$  as merely 1 or 5, thus limiting their performance.


\section{Conclusion}

In this paper, we propose LMKE, an effective and efficient method to adopt language models as  knowledge embeddings. 
Motivated by the inability of structure-based KEs to well represent long-tail entities, LMKE leverages textual descriptions and learns embeddings of entities and relations in the same space as word tokens.
It solves the restrictive demand of prior description-based approaches.   
A contrastive learning framework is further proposed that allows zero-cost negative sampling and  significantly downgrades time complexity in both training and evaluation. 
Results of extensive experiments show that our methods achieve state-of-the-art performance on various benchmarks, especially for long-tail entities. 


In the future, we plan to explore the effectiveness of more \wxt{advanced} contrastive learning approaches 
in description-based KEs. \wxt{We are also interested in language models' capacity of modeling composition patterns in KGs.}


\section*{Acknowledgements}
This work was supported by National Key Research and Development Project
(No.2020AAA0109302), Shanghai Science and Technology Innovation Action Plan (No.19511120400), Shanghai Municipal Science
and Technology Major Project (No.2021SHZDZX0103), China Postdoctoral Science Foundation (No. 2020M681173 and No. 2021T140124), and National Natural Science Foundation of China (Grant No. 62102095).

\bibliographystyle{named}
\bibliography{ijcai22}

\end{document}